\documentclass[10pt,twocolumn,letterpaper]{article}

\usepackage[accsupp]{axessibility}  % Improves PDF readability for those with disabilities.

\usepackage{cvpr}              % To produce the CAMERA-READY version

\usepackage{amsmath}
\usepackage{mdframed}
\usepackage{makecell}
\usepackage{url}
\usepackage[most]{tcolorbox}
\usepackage{enumitem}
\usepackage{multirow}
\usepackage{float}
\usepackage{tabularx}
\usepackage{colortbl}
\usepackage{diagbox}  % Include this in your preamble
\usepackage{booktabs} % Optional: for nicer lines
\usepackage{amssymb}

\definecolor{cvprblue}{rgb}{0.21,0.49,0.74}
\usepackage[pagebackref,breaklinks,colorlinks,allcolors=cvprblue]{hyperref}

\def\paperID{20914} % *** Enter the Paper ID here
\def\confName{CVPR}
\def\confYear{2026}

\title{Benchmarking Vision-Language Models under Contradictory Virtual Content Attacks in Augmented Reality}

\author{
Yanming Xiu, Zhengyuan Jiang, Neil Zhenqiang Gong, Maria Gorlatova\\
Duke University\\
{\tt\small \{yanming.xiu, zhengyuan.jiang, neil.gong, maria.gorlatova\}@duke.edu}
}

\begin{document}
\maketitle

\begin{abstract}
Augmented reality (AR) has rapidly expanded over the past decade. As AR becomes increasingly integrated into daily life, its security and reliability emerge as critical challenges. Among various threats, contradictory virtual content attacks, where malicious or inconsistent virtual elements are introduced into the user’s view, pose a unique risk by misleading users, creating semantic confusion, or delivering harmful information. In this work, we systematically model such attacks and present ContrAR, a novel benchmark for evaluating the robustness of vision-language models (VLMs) against virtual content manipulation and contradiction in AR. ContrAR contains 312 real-world AR videos validated by 10 human participants. We further benchmark 11 VLMs, including both commercial and open-source models. Experimental results reveal that while current VLMs exhibit reasonable understanding of contradictory virtual content, room still remains for improvement in detecting and reasoning about adversarial content manipulations in AR environments. Moreover, balancing detection accuracy and latency remains challenging.
\end{abstract}
\vspace{-0.5cm}

\section{Introduction}
\label{sec:intro}

\begin{figure}[t]
\includegraphics[width=1\linewidth]{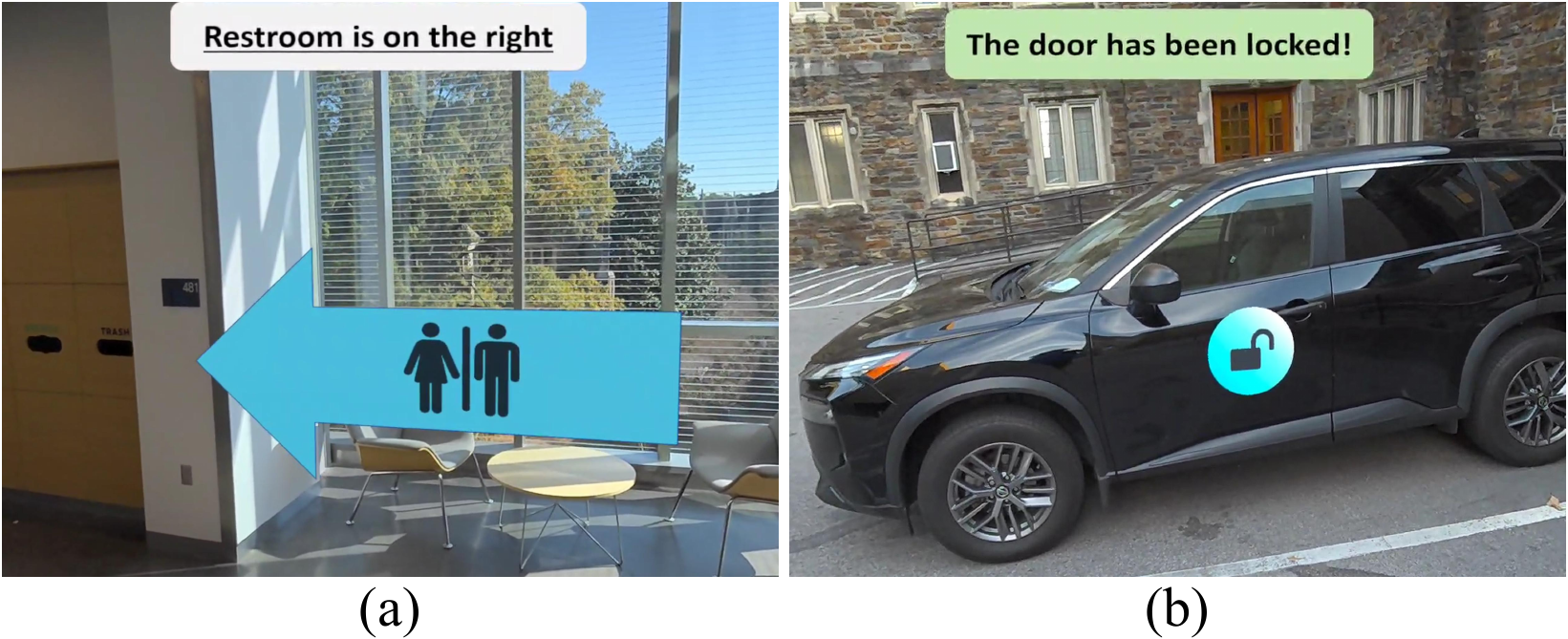}
\centering
\vspace{-0.8cm}
% \caption{Examples of contradictory virtual content in augmented reality. (a) The text instruction ``Restroom is on the right" conflicts with the directional arrow pointing left; (b) The virtual icon indicates that the car door is unlocked, while the text instruction claims ``The door has been locked!". These semantic inconsistencies illustrate contradictory virtual content attacks, which may confuse AR users or provide misleading information.}
\caption{Examples of contradictory virtual content in augmented reality. (a) The text instruction ``Restroom is on the right" conflicts with the directional arrow pointing left; (b) The virtual icon indicates that the car door is unlocked, while the text instruction claims ``The door has been locked!"}
\label{fig:cvc example}
\vspace{-0.8cm}
\end{figure}

Augmented Reality (AR) overlays digital information onto the physical world, providing users with additional information and enabling users' intuitive interactions with surrounding environments. Modern AR experiences often integrate virtual content from multiple sources, such as the system UI and different commercial applications. As AR ecosystems grow increasingly open and interconnected, these virtual elements may either unintentionally or maliciously introduce visual attacks~\cite{attack01, attack02, VIM03, viddar, VIM01, xiu2025demonstrating} in AR scenes. One of the major concerns is the contradiction attack~\cite{contradiction01, contradiction02} caused by virtual content in AR scenes, where virtual content presents inconsistent or conflicting semantics within the same scene. For example, in Fig.~\ref{fig:cvc example}, the contradictory virtual content can potentially lead users to incorrect destinations or raise property safety concerns. 

While such contradictory virtual content can lead to negative user experience and consequences, detecting them is challenging because it requires semantic-level reasoning beyond low-level feature analysis. The system must not only recognize the virtual content and their attributes, but also infer relationships, consistency, and intent among coexisting virtual contents. Previous efforts on virtual content analysis in AR have primarily focused on quality assessment, such as analyzing alignment accuracy, lighting consistency, or perceptual realism, without modeling high-level semantic consistency among virtual objects. These approaches can quantify rendering or placement quality but fail to capture when virtual elements convey conflicting meanings. 

Conversely, a rich body of work on semantic reasoning, spanning visual question answering, scene understanding, and multimodal reasoning, has demonstrated impressive progress in general computer-vision contexts. Recent advances in vision-language models (VLMs) have demonstrated remarkable performance on a wide range of semantic reasoning tasks that combine visual perception and linguistic understanding. Models such as GPT~\cite{GPT5}, Gemini~\cite{gemini2.5} and Grok~\cite{Grok4} can perform multimodal question answering, captioning, and reasoning about object relations. These capabilities make VLMs promising candidates for understanding and detecting semantic contradictions in AR scenes, where both visual and textual information coexist and interact. However, the methods and models are usually tested outside the AR setting, so it remains unclear whether they can still robustly interpret or detect semantic contradictions in complex, mixed-reality scenes.

Motivated by this observation, we aim to systematically study contradictory virtual content attacks in AR and evaluate the capability of modern VLMs to recognize such semantic conflicts. To this end, we propose ContrAR, a new benchmark that models this type of adversarial manipulation and provides a standardized dataset for evaluating VLM performance under contradictory virtual content attacks. ContrAR is built from real-world AR scenes and includes human validation to ensure label annotation quality. Our main contributions are as follows:

\begin{itemize}
    \item We formally define the contradictory virtual content attack through a threat model tailored to commercial AR systems. The model specifies the attacker's purpose, knowledge, and capabilities under realistic constraints, as well as the attack detection system's accessible information and operational limits. Our threat model provides a foundation for analyzing semantic-level AR threats.
    \item We propose \textit{ContrAR}, the first benchmark dataset that systematically represents contradictory virtual content attacks in AR, containing 312 real-world AR videos collected by a commercial head-mounted device (HMD), Meta Quest 3. The labels of the videos are validated by 10 human subjects. The dataset is available on GitHub.\footnote{\label{dataset}\url{https://github.com/YM-Xiu/ContrAR-Dataset}}
    \item We benchmark 11 VLMs, including commercial models and open-source models, and analyze their strengths and weaknesses in detecting semantic contradictions in AR content. Our findings reveal that while these models demonstrate a reasonable understanding of contradictory virtual content in AR, their performance still leaves significant room for improvement, and achieving a balance between accuracy and latency remains an open challenge.
\end{itemize}

The remainder of this paper is organized as follows:
Sec.~\ref{sec:related} reviews prior work related to contradictory content attack, virtual content analysis, and AR scene understanding.
Sec.~\ref{sec:threat_model} defines the threat model and attacker assumptions that guide our benchmark design.
Sec.~\ref{sec:dataset} introduces the construction and annotation of the proposed ContrAR dataset, while Sec.~\ref{sec:exp} presents the experiment setup, reports the results, and provides analysis on model performance and latency trade-offs.
Finally, Sec.~\ref{sec:limitation} discusses the limitations and future directions, and Sec.~\ref{sec:conclusion} concludes the paper.

\section{Related Work}
\label{sec:related}

\subsection{Contradictory Content Attack}

Semantically contradictory information has long been recognized as a safety and reliability concern in natural language understanding and human-computer interaction.
Early work on contradiction detection in text defined types of contradictory statements and developed corpora for sentence-pair inference~\cite{contra01}. 
Recent work, such as BoardgameQA~\cite{contradiction01}, introduces textual benchmarks that test models' reasoning over contradictory information in language, demonstrating that large models still perform inconsistently when evidence conflicts. 
Pan et al.~\cite{contradiction02} studied the effect of injecting contradictory misinformation into open-domain question-answering corpus and found that current models are vulnerable to small amounts of semantic conflict in retrieved evidence. 
% Gärtner~\cite{contra03}  proposed ALICE, a hybrid logic-LLM system for detecting semantic contradictions in formal engineering requirements. 
However, these efforts remain confined to the text domain and do not extend to visual or multimodal contexts.

In computer vision research, exploration of semantic contradictions mainly focuses on the semantic information in images. Pedziwiatr et al.~\cite{contra02}  examined how semantic contradictions between objects and scenes influence visual attention, showing that contradictory objects attract more fixations but are perceived as less meaningful. More recently, multimodal benchmarks such as MMIR~\cite{multimodalinconsistency} have begun investigating visual-textual inconsistencies in layout-rich documents and presentations. However, most existing related work remains limited to still-image inconsistencies and thus does not address the unique challenges posed by the dynamic mixed-reality environments, where multiple virtual elements must be interpreted in the context of the real world continuously. Also, previous visual inconsistency studies mainly focus on anomaly detection in natural images or UI interfaces, rather than purposefully contradictory virtual content embedded in AR scenes. This gap motivates our definition of contradictory virtual content attacks in AR and the creation of the ContrAR benchmark to systematically evaluate them.

\subsection{Virtual Content Analysis in AR}

Assessing and evaluating the quality of virtual content has long been an important topic in AR research. A significant amount of work on virtual content quality assessment primarily examines low-level visual or spatial metrics, such as rendering fidelity, lighting realism, depth alignment, and registration accuracy~\cite{ARassessshadow01, ARassessmentlighting01, ARassessmisalign01, ARassessphysics01, ARassessmisalign02, ARIQA01, chen2025neurosymbolic, DiverseARplus}. These approaches ensure geometric or perceptual coherence but lack awareness of the semantic-level information conveyed by the virtual elements and their relationships within the AR scene. As a result, they cannot evaluate whether virtual content delivers information that is meaningful, consistent, and trustworthy to the user.

Another related direction is AR scene understanding, which seeks to interpret the semantic structure of the AR environment, including object recognition, depth estimation, and spatial reasoning. This direction has been applied in fields such as robotics~\cite{sceneunderstanding01}, surgery~\cite{sceneunderstanding02}, and assembly~\cite{sceneunderstanding03}. More recently, several studies have applied generative AI models, especially VLMs, to enhance semantic reasoning in AR tasks, including virtual content description~\cite{Genaixrws} and virtual object placement~\cite{xair, octo}. While these works advance a holistic understanding of AR environments, they are developed under benign conditions and do not consider adversarial or malicious content. Some work has further extended the application of VLMs to AR safety research toward identifying and mitigating detrimental or adversarial virtual content, such as obstruction detection~\cite{viddar}, visual information manipulation detection~\cite{vimsense}, privacy protection~\cite{chen2024securing}, and user cognition protection~\cite{ARcognitiveattack}. These systems demonstrate that multimodal reasoning can support high-level interpretation of AR scenes and improve safety awareness. However, all of them provide system-level solutions tailored to applications rather than a unified benchmark for model evaluation. In our work, we aim to propose a standardized benchmark that isolates the semantic contradiction aspect of AR content and enables reproducible evaluation of VLM reasoning under adversarial conditions.

\section{Threat Model}
\label{sec:threat_model}

We start our work by establishing a threat model, which defines the basic assumptions, the attacker's purposes, knowledge, and capabilities, as well as the capabilities of the attack detection system. This model serves as the conceptual foundation for understanding contradictory virtual content attacks in AR and for designing the subsequent benchmark and evaluation.

\subsection{Basic Assumptions}
% \myparatight{Basic Assumptions}

We consider an AR environment deployed on an HMD. The device hardware, including the camera and spatial tracking sensors, is assumed to function correctly and to provide authentic data. The operating system (OS) and compositor are trusted and are responsible for isolating user-level applications and correctly compositing their outputs. Applications are sandboxed and cannot gain unauthorized access to the memory or internal states of other apps. Cloud-based services, such as semantic reasoning, are also considered trustworthy, and all communication with them is assumed to be encrypted and uncompromised. The user is benign and interacts with the AR system in good faith. Under these assumptions, any security risk arises from malicious or misleading virtual content generated at the application level, rather than from hardware or system compromise.

\subsection{Attacker Purpose}

The attacker seeks to manipulate the semantics of virtual content shown to the user rather than to directly disrupt system operation. The primary objectives include: (1) misleading the user by displaying virtual elements that convey conflicting or false information (e.g., presenting contradictory navigation cues that lead a user to the wrong destination, thereby reducing task efficiency and causing delays); (2) inducing semantic confusion through conflicting messages or object-label pairs that make correct interpretation difficult; and (3) degrading user safety, trust, or decision quality by introducing deceptive cues that produce harmful outcomes, such as the examples shown in Fig.~\ref{fig:dataset}. Collectively, these behaviors constitute a contradictory virtual content attack, in which multiple virtual elements present mutually inconsistent semantics within the same AR scene.

% , which presenting an incorrect foreign-exchange rate that prompts an unreasonable financial action, or falsified status indicators that cause unsafe operation).

\subsection{Attacker Knowledge}

We assume a gray-box threat setting. The attacker application operates with the same privileges as a legitimate AR application and can access only information that is publicly exposed through standard APIs. Specifically, the attacker can observe the composited scene visible to the user, including passthrough camera frames, depth information, and virtual content rendered by its own application, when such access is permitted by the platform. The attacker can also possess general knowledge of the victim application’s intended functionality and of the platform’s rendering and composition pipeline. However, the attacker has no privileged insight into the operating system or hardware firmware, nor can it obtain private data or rendering parameters from other applications. Cloud-based services used by the platform or by legitimate applications are treated as trusted and non-compromisable under this model.

\subsection{Attacker Capability}

Given the above knowledge, the attacker application can create and render arbitrary virtual objects, text, or overlays within its own application using legitimate rendering APIs. It can control what content to display, where to place it in the scene, and when to present or remove it, enabling semantic manipulations such as contradictory labels, conflicting indicators, or temporally alternating cues. The attacker may utilize both local computation and external (trusted) cloud resources to generate or update these assets. However, it cannot alter virtual content rendered by other applications, tamper with sensor data, modify the compositor’s blending process, intercept inter-application communication, or escalate privileges to access protected OS components. Hardware spoofing and denial-of-service attacks are out of our scope.

\subsection{Attack Detection System Capability}

Similar to the attacker application, the attack detection system also runs as a user-level process observing the same composited frames visible to the user. It has view-only access and cannot query per-application render lists or use privileged APIs; consequently, it cannot attribute a specific virtual element to its source application. The detector employs multimodal reasoning components, especially VLMs, to analyze relationships among virtual elements and identify inconsistencies. Its goal is to detect contradictory or manipulative virtual content in real time using only visual evidence. Cloud-based VLM reasoning services used by the detector are trusted for correctness and integrity.

\subsection{Scope of Attack}

This work focuses on semantic-level attacks that alter the meaning or relationships among virtual elements without compromising the underlying hardware or OS. Physical tampering, sensor spoofing, or system-level privilege escalation are excluded. An attack is considered successful if contradictory virtual content is successfully introduced into the user’s view and remains undetected by the detector within a reasonable temporal window. In conclusion, all three entities (the victim application, the attacker application, and the detection system) operate with ordinary application-level privileges, and the threat model is confined to semantic information manipulation and contradiction rather than any form of system-level compromise.

\section{ContrAR Dataset}
\label{sec:dataset}

Building on the threat model above, we construct a dataset, ContrAR, which systematically represents contradictory virtual content attacks across multiple practical AR applications. Our design goals are realism, diversity of attack modes, and reproducible annotation.

\subsection{Contradictory Virtual Content Attack Design}

Following the threat model introduced in Section~\ref{sec:threat_model}, we formally define the structure of a \textit{contradictory virtual content attack}. In such attacks, the attacker introduces virtual elements whose semantics conflict with each other, thereby misleading the user. Importantly, the attack is defined purely at the semantic level: the attack detector does not need to and cannot identify which element originates from the attacker or the victim application, but only whether a contradiction exists within the overall composited view.

Formally, an AR scene may contain multiple virtual elements that together form a set of $n$ virtual contents:

\vspace{-0.3cm}
\begin{equation}
\mathcal{C} = \{ c_1, c_2, \ldots, c_n \},
\end{equation}
\vspace{-0.4cm}

where each $c_i$ represents a distinct virtual content, which can be an object, a label, or text visible to the user. Each content conveys a piece of semantic information $I(c_i)$ describing its meaning or intent in the scene context. The collection of all such semantics is represented as
\begin{equation}
\mathcal{I} = \{ I(c_1), I(c_2), \ldots, I(c_n) \}.
\end{equation}

A scene is considered to exhibit a \textit{contradictory virtual content attack} if at least two pieces of information in $\mathcal{I}$ are semantically inconsistent, that is:
\begin{equation}
\exists\,(I(c_i), I(c_j)) \in \mathcal{I}^2,\, i \neq j, \text{ such that } I(c_i)\,\perp\, I(c_j),
\end{equation}
where the relation $\perp$ denotes semantic contradiction between two statements. Intuitively, this means that the corresponding virtual elements communicate mutually exclusive or logically incompatible information, such as an arrow pointing to left versus text ``Turn Right," or an unlocked sign versus text `the door has been locked", as Fig. ~\ref{fig:cvc example} shows.

If no such contradictory pair exists, i.e.,

\vspace{-0.5cm}
\begin{equation}
\forall\,(I(c_i), I(c_j)) \in \mathcal{I}^2,\, i \neq j,\, I(c_i) \not\!\perp I(c_j),
\end{equation}
\vspace{-0.5cm}

the scene is labeled as \textit{non-contradictory}. We then define the ground-truth label for a video $V$ as

\vspace{-0.2cm}
\begin{equation}
C(V) =
\begin{cases}
1, & \text{if } \exists\, I(c_i)\perp I(c_j),\\
0, & \text{otherwise.}
\end{cases}
\end{equation}
\vspace{-0.2cm}

This formalization provides a logical criterion for identifying contradictory virtual content attacks. It also aligns with our detector’s design objective: to determine whether a scene contains contradictory content, rather than attributing specific content to attacker or victim applications.

\begin{figure}[t]
\includegraphics[width=1\linewidth]{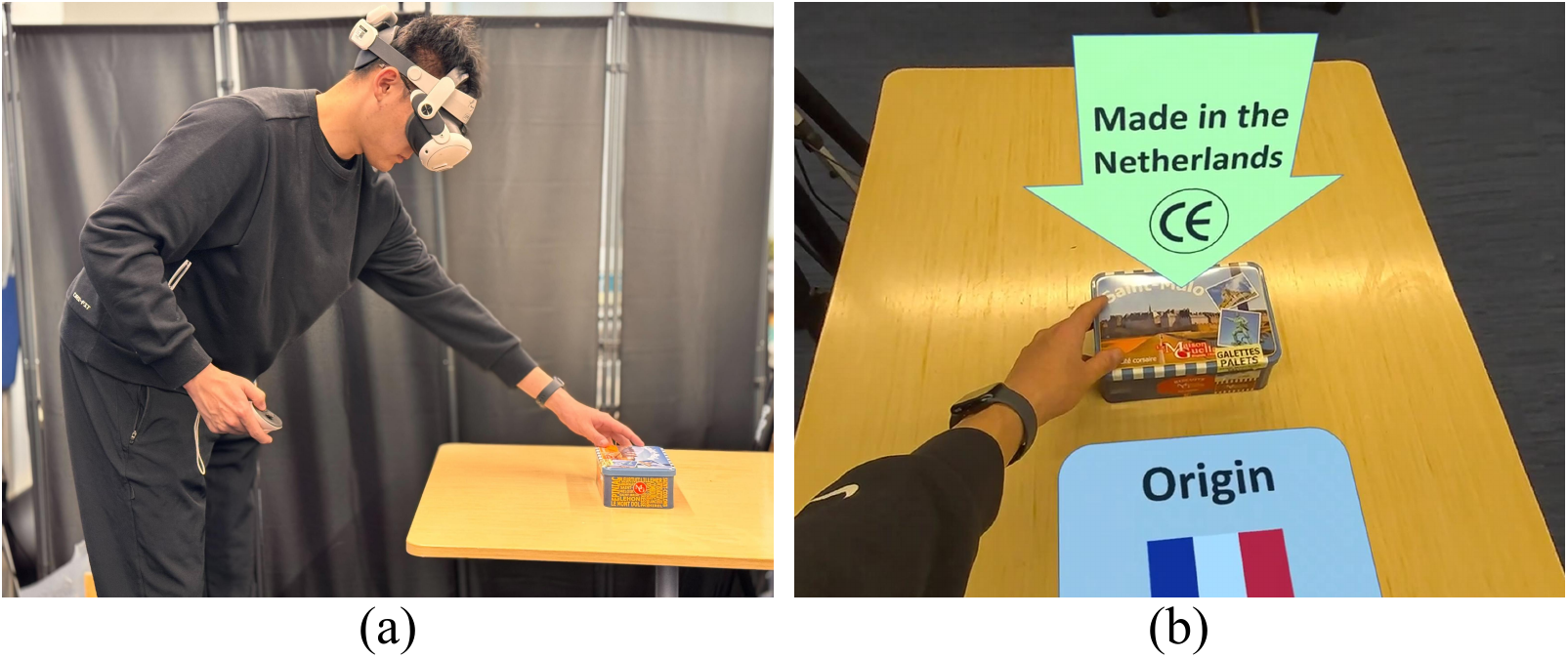}
\centering
\vspace{-0.7cm}
\caption{Demonstration of the dataset collection process. (a) AR user’s view; (b) bystander’s view.}
\label{fig:data collection}
\vspace{-0.6cm}
\end{figure}

\begin{figure*}[t]
\includegraphics[width=0.9\linewidth]{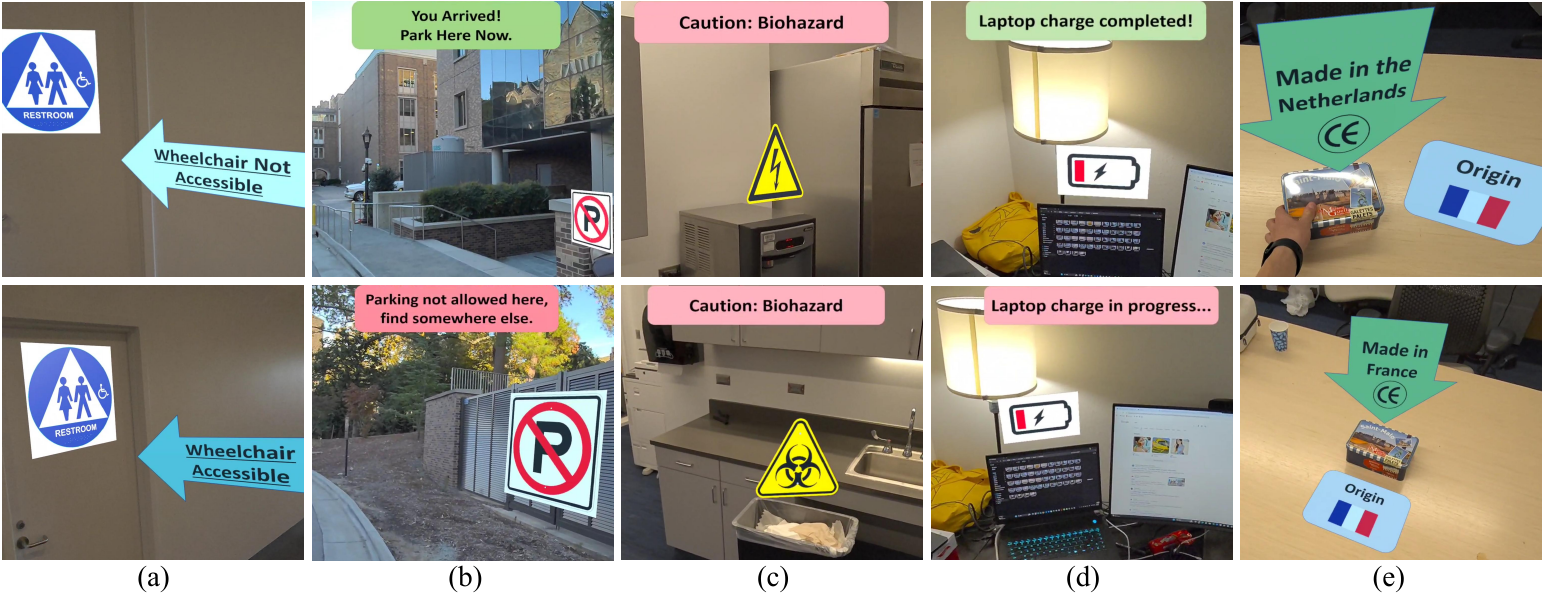}
\centering
\vspace{-0.3cm}
% \caption{Samples frames from the ContrAR dataset. Each column shows an AR application. The top row shows the contradictory examples. From left to right: (a) Indoor navigation: the text on the virtual arrow indicates the restroom is not accessible, which contradicts with the disabled sign on the virtual restroom sign, potentially avoid the disabled from using the restroom; (b) Outdoor navigation: the text in the virtual dialog box instructs the user to park here, which is contradictory to the information provided by a virtual ``No Parking" sign, leading to potential fines; (c) Safety inspection: the text in the virtual dialog box ``Caution: Biohazard" is contradictory with the virtual sign which stands for ``Caution: High Voltage", which may confuse the user about the safety regulations in this area; (d) smart apartment: the text in the virtual dialog box indicates the laptop is fully charged, which contradicts the virtual ``low battery" sign, which may lead to an early stop of charging; (e) smart retail: the text on the virtual arrow tells the user the box is made in the Netherlands, which contradicts to the label next to the box saying it is from France, which may lead to non-voluntary purchasing behavior. The bottom row shows the non-contradictory examples, where the information provided by different virtual content aligned with each other.}
\caption{Sample frames from the ContrAR dataset. Each column shows an AR application. The top row shows the contradictory examples. From left to right: (a) Indoor navigation: the text on the virtual arrow indicates the restroom is not accessible, which contradicts the disabled sign on the virtual restroom sign; (b) Outdoor navigation: the text in the virtual dialog box instructs the user to park here, which is contradictory to the information provided by a virtual ``No Parking" sign; (c) Safety inspection: the text in the virtual dialog box ``Caution: Biohazard" is contradictory with the virtual sign which stands for ``Caution: High Voltage"; (d) Smart apartment: the text in the virtual dialog box indicates the laptop is fully charged, which contradicts the virtual ``low battery" sign; (e) Smart retail: the text on the virtual arrow tells the user the box is made in the Netherlands, which contradicts the label next to the box saying it is from France. The bottom row shows the non-contradictory examples, where the information provided by different virtual content aligns with each other.}
\label{fig:dataset}
\vspace{-0.5cm}
\end{figure*}

% \begin{table*}[t]
% \caption{Data distribution of ContrAR dataset by victim AR apps and attack labels.}
% \footnotesize
% \vspace{-0.2cm}
% \centering
% {
% \renewcommand{\arraystretch}{0.8}
% \begin{tabular}{c|c|c|c|c|c|c}
% \toprule
% \textbf{Victim AR Application} &
% \makecell{Indoor Navigation} &
% \makecell{Outdoor Navigation} &
% \makecell{Safety Inspection} &
% \makecell{Smart Apartment} &
% \makecell{Smart Retail} &
% \makecell{Total Number}\\
% \midrule
% \textbf{Contradictory}     & 27 & 30 & 21 & 45 & 33 & 156\\
% \textbf{Non-Contradictory}   & 27 & 30 & 21 & 45 & 33 & 156\\
% \midrule
% \textbf{Total Number}        & 54 & 60 & 42 & 90 & 66 & 312\\
% \bottomrule
% \end{tabular}
% }
% \label{tab:dataset_distribution}
% \vspace{-0.4cm}
% \end{table*}

\subsection{Dataset Collection Pipeline}

To capture realistic attacks, we implement five representative victim AR application cases on Meta Quest 3 that reflect common, safety-relevant use cases: indoor navigation (IN), outdoor navigation (ON), safety inspection (SI), smart apartment (SA), and smart retail (SR). Each application embodies plausible user-facing functionality and therefore provides natural opportunities for semantic manipulation. For each application, contradictory attack patterns were manually designed through structured brainstorming with three researchers with expertise in AR design, focusing on the real-world background and semantically conflicting virtual cues. We design attacker behaviors for each scene of each app to mirror the attacker purposes in our threat model (misleading navigation cues, contradictory status indicators, etc.), enabling evaluation of detection under diverse threat instances.

For convenience and reproducibility in data collection, a single unified app is used to simulate all five AR use cases and also act alternately as the victim app and the attacker app during recording.  This design choice aligns with our assumption about the attack detector, which operates at the semantic level and does not possess the capability to distinguish between content generated by different applications.  In each recorded session, the app places the benign virtual content based on the real-world context and, depending on the experimental condition, an additional set of hypothetical attacker-generated virtual elements that either introduce a semantic contradiction or do not. All of the virtual content is precisely placed at the manually designated location with the help of Meta Quest 3's built-in spatial tracking functionality. Each resulting video therefore contains the composited scene that a real user would observe, and is labeled as contradictory or non-contradictory.

During data collection, we used a Meta Quest 3 headset as the AR platform. The AR application was developed in Unity 6000.1.7f1, employing ARCore for spatial tracking. The Meta SDK was used to access the main camera stream, and all recordings were captured using the headset’s built-in screen-recording functionality to ensure fidelity to the user’s actual view. An example figure showing our video collection process is shown in Fig. \ref{fig:data collection}.

\begin{table}[t]
\caption{Data distribution of ContrAR dataset by victim AR apps and attack labels.}
\footnotesize
\vspace{-0.2cm}
\centering
{
\renewcommand{\arraystretch}{0.8}
\begin{tabular}{c|c|c|c|c|c|c}
\toprule
\textbf{Victim AR Application} &
\makecell{IN} &
\makecell{ON} &
\makecell{SI} &
\makecell{SA} &
\makecell{SR} &
\makecell{Total}\\
\midrule
\textbf{Contradictory}     & 27 & 30 & 21 & 45 & 33 & 156\\
\textbf{Non-Contradictory}   & 27 & 30 & 21 & 45 & 33 & 156\\
\midrule
\textbf{Total Number}        & 54 & 60 & 42 & 90 & 66 & 312\\
\bottomrule
\end{tabular}
}
\label{tab:dataset_distribution}
\vspace{-0.55cm}
\end{table}

\subsection{Dataset Composition}

With the data collection pipelines, we develop the ContrAR dataset. ContrAR consists of 312 videos, each representing a unique AR scenario. To ensure a balanced composition and avoid bias, we collected a strict 1:1 ratio of contradictory (positive) and non-contradictory (negative) video samples. Among all the videos, 90 videos contain only textual virtual content, while 222 videos contain both visual and textual virtual content. Each video in the dataset has a resolution of 1920×1080 pixels and a duration ranging from 5 to 15 seconds. The videos had a framerate of 30 FPS. The full dataset composition is shown in Tab.~\ref{tab:dataset_distribution}. Some of the data examples, together with the attack labels, are shown in Fig.~\ref{fig:dataset}. The full dataset, including video and attack labels, is publicly released on Github.

\subsection{User-Based Data Validation}

\begin{figure}[t]
\includegraphics[width=1\linewidth]{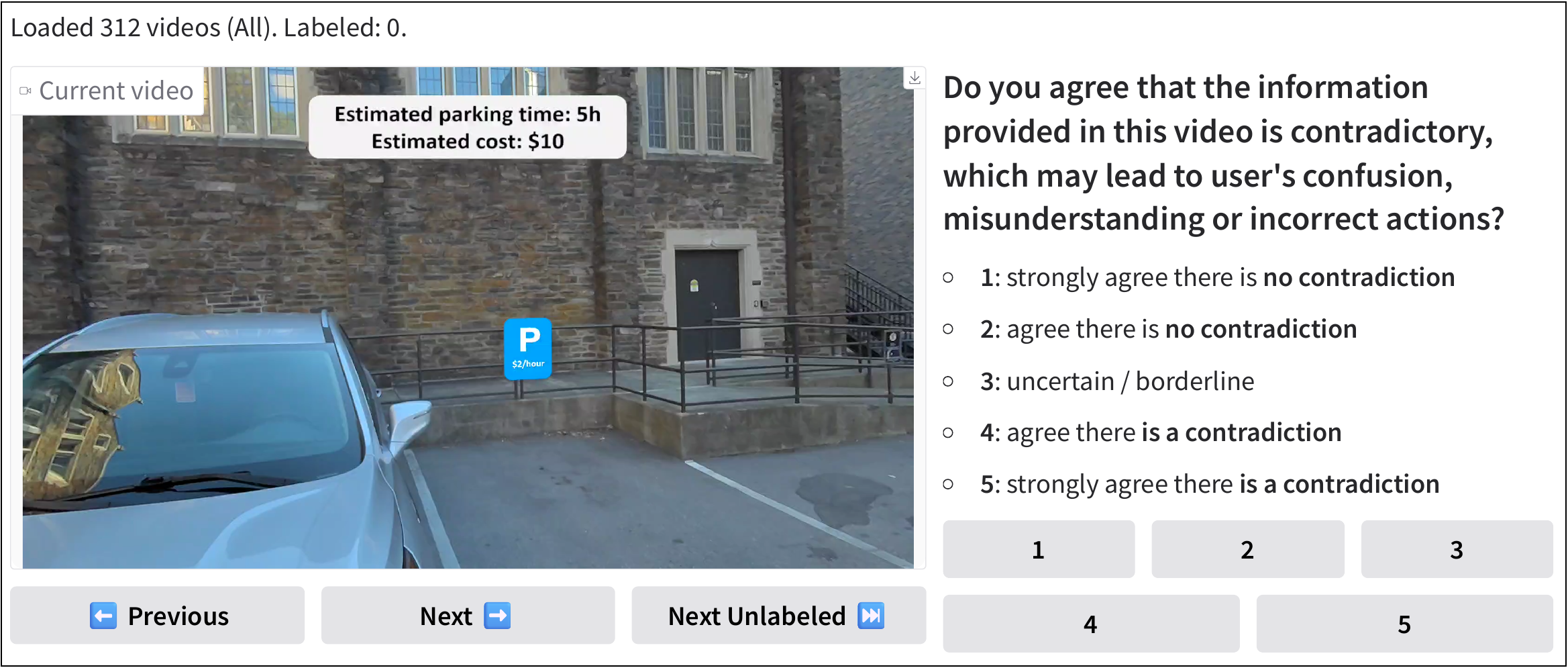}
\centering
\vspace{-0.6cm}
\caption{Interface used in the user-based labeling task for evaluating contradictory virtual content in AR videos.}
\label{fig:UI}
\vspace{-0.3cm}
\end{figure}

With the collected videos, we manually assigned each sample a label of either ``\textit{Contradictory}" or ``\textit{Non-contradictory}." To evaluate the reliability of these labels, we conducted an IRB-approved, user-based data validation study. A group of 10 participants was recruited to review the entire dataset and independently determine whether each video contained contradictory virtual content. For each video, participants rated their level of agreement with the question ``Do you agree that the information provided in this video is contradictory, which may lead to users' confusion, misunderstanding, or incorrect actions?" on a 5-point Likert scale~\cite{likert}, ranging from 1 (strongly disagree) to 5 (strongly agree). 
Although the question refers to possible confusion or misunderstanding, these effects were not used as labeling criteria. Instead, they were included only to illustrate the potential consequences of contradictory content and to make the notion of contradiction more concrete for participants.
For non-contradictory samples, where disagreement indicates a correct judgment, the scores were inverted by subtracting each value from 6, ensuring that higher scores consistently reflected stronger agreement with the ground-truth label. The user interface used for this validation process is built with Gradio~\cite{gradio}, as illustrated in Fig.~\ref{fig:UI}.

\begin{figure}[t]
\includegraphics[width=0.95\linewidth]{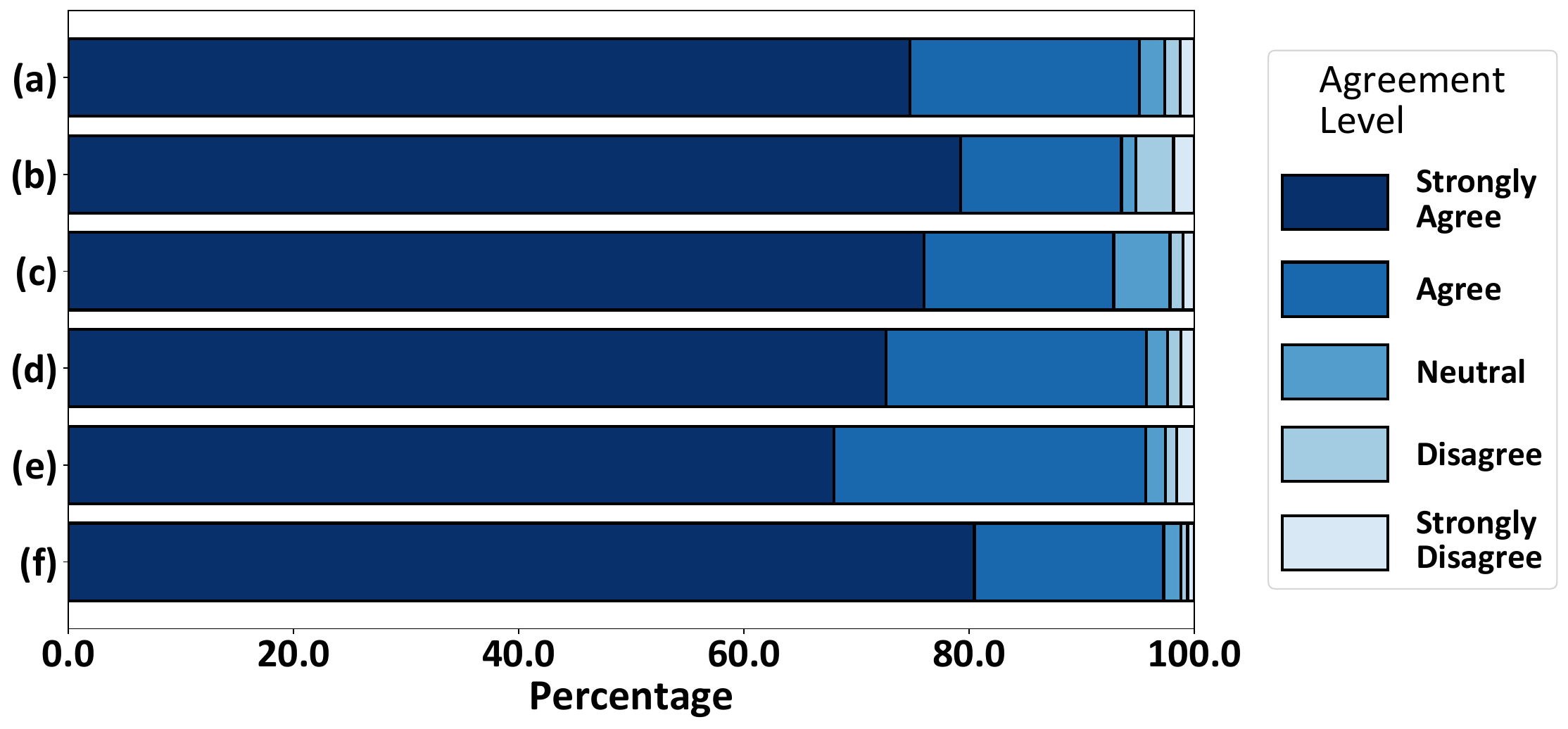}
\centering
\vspace{-0.3cm}
\caption{User agreement with contradictory virtual content attack labels in the ContrAR dataset. (a) The overall distribution of Likert-scale responses. (b)-(f) Likert responses for all five AR application scenes: (b) indoor navigation, (c) outdoor navigation, (d) safety inspection, (e) smart apartment, (f) smart retail.}
\label{fig:likert}
\vspace{-0.6cm}
\end{figure}

The agreement score distribution is shown in Fig.~\ref{fig:likert}, and the average score across all videos is 4.66, demonstrating strong consistency between the dataset annotations and human judgments. These user validation results confirm the perceptual reliability of ContrAR and indicate that the designed contradictory virtual content attacks are generally recognized and agreed upon by human participants.

\vspace{-0.3cm}

\section{Experiment and Results}
\label{sec:exp}

\subsection{Experiment Design}

With the constructed dataset, we conducted a comprehensive evaluation of multiple commercial VLMs, including GPT~\cite{GPT4V, GPT5}, Gemini~\cite{geminiteam2024geminifamilyhighlycapable, gemini2.5}, Grok~\cite{Grok4}, and Claude~\cite{anthropic2025claudesonnet45, anthropic2025claudehaiku45}, as well as two open-source models, Qwen-2.5-VL-72B and Qwen-2.5-VL-7B. In addition to these image-based VLMs, we include a text-only LL baseline that assesses contradiction purely from extracted textual overlays. For each video, we applied an OCR engine, EasyOCR, to the same selected frame(s) and aggregated all detected text snippets into a structured list. This text-only representation was then provided to GPT-4o, which was prompted to judge whether the extracted textual information exhibited semantic contradiction.

\begin{figure}[t]
\includegraphics[width=0.85\linewidth]{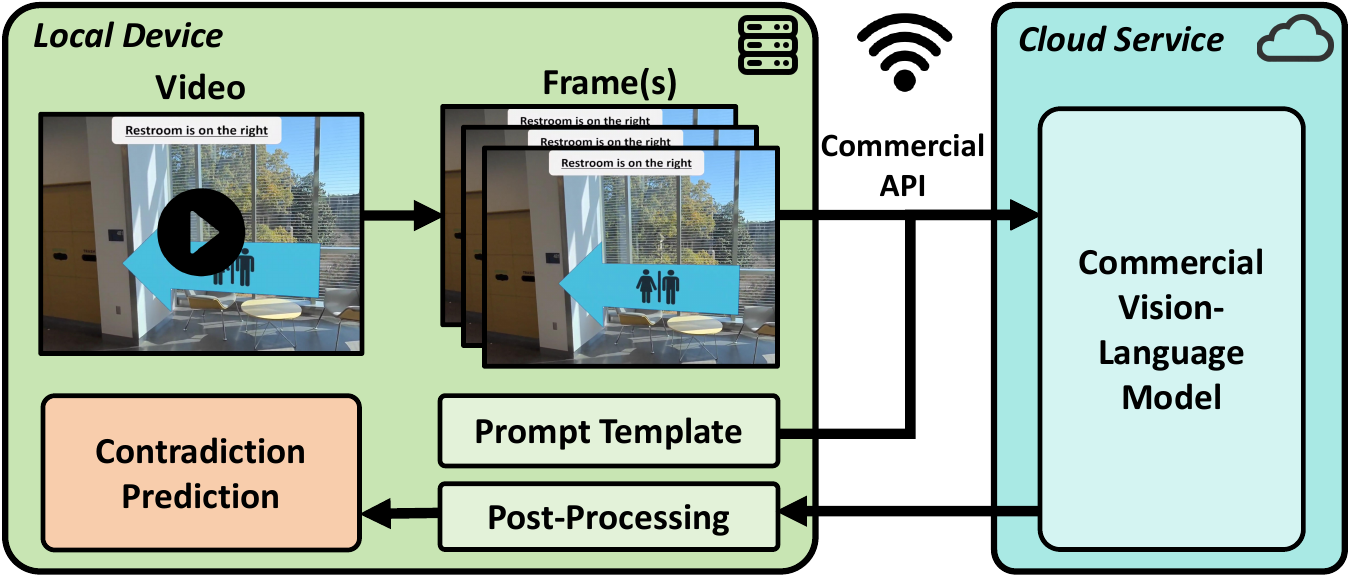}
\centering
\vspace{-0.2cm}
\caption{Pipeline for evaluating VLMs on the ContrAR.}
\label{fig:diag}
\vspace{-0.6cm}
\end{figure}

\begin{table*}[ht]
\caption{Detection accuracy and latency of different detection methods on ContrAR benchmark.}
\footnotesize
\vspace{-0.2cm}
\renewcommand{\arraystretch}{0.8}
\setlength{\tabcolsep}{5pt}
\centering
\begin{tabular}{c|c|ccccc|c|c}
\toprule
\multirow{3}{*}{\makecell{Detection Method}} & 
\multirow{3}{*}{\makecell{Prompting \\ Method}} & \multicolumn{6}{c|}{Contradictory Virtual Content Attack Detection Accuracy (\%)} & \multirow{3}{*}{\makecell{Detection Latency \\ (second)}} \\
\cmidrule(lr){3-8}
% & & Char Repl. & Phra Repl. & Phra Obfu. & Phra Extr. & Patt Repl. & Patt Obfu. & Patt Extr. & \textbf{Overall} \\
& & 
\makecell{Indoor \\ Navigation} & 
\makecell{Outdoor \\ Navigation} & 
\makecell{Safety \\ Inspection} & 
\makecell{Smart \\ Apartment} & 
\makecell{Smart \\ Retail} & 
\makecell{Overall}\\
\midrule
\multirow{2}{*}{\textbf{GPT-5}} 
& Single Frame         & 81.48 & \textbf{91.67} & 80.95 & \textbf{94.44} & \textbf{86.36} & \textbf{88.14} & 19.29 \\
& Multiple Frames  & 83.33 & 88.33 & 78.57 & 92.22 & 80.30 & 85.58 & 23.78 \\
\midrule
\multirow{2}{*}{\textbf{GPT-4.1}} 
& Single Frame         & 79.63 & 86.67 & \textbf{90.48} & 84.44 & 71.21 & 82.05 & 11.47 \\
& Multiple Frames  & 85.19 & 85.00 & 88.10 & 93.33 & 78.79 & 86.54 & 16.61 \\
\midrule
\multirow{2}{*}{\textbf{GPT-4o}} 
& Single Frame         & 83.33 & 86.67 & 71.43 & 77.78 & 75.76 & 79.17 & 5.92 \\
& Multiple Frames  & \textbf{88.89} & 90.00 & 80.95 & 86.67 & 75.76 & 84.62 & 7.26 \\
\midrule
\multirow{2}{*}{\textbf{Gemini-2.5-Pro}} 
& Single Frame         & 75.93 & 90.00 & 83.33 & 86.67 & 81.82 & 83.97 & 14.29 \\
& Multiple Frames  & 83.33 & 81.67 & 66.67 & 82.22 & 65.15 & 76.60 & 15.56 \\
\midrule
\multirow{2}{*}{\textbf{Gemini-2.5-Flash}} 
& Single Frame         & 85.19 & 80.00 & 69.05 & 88.89 & 69.70 & 79.81 & 9.90 \\
& Multiple Frames  & 75.93 & 76.67 & 59.52 & 76.67 & 63.64 & 71.47 & 10.58 \\
\midrule
\multirow{2}{*}{\textbf{Grok-4}} 
& Single Frame         & 75.93 & 75.00 & 64.29 & 61.11 & 68.18 & 68.27 & 27.76 \\
& Multiple Frames  & 64.81 & 70.00 & 64.29 & 57.78 & 69.70 & 64.74 & 49.09 \\
\midrule
\multirow{2}{*}{\textbf{Grok-2-Vision}} 
& Single Frame         & 74.07 & 70.00 & 61.90 & 57.78 & 63.64 & 64.74 & 4.66 \\
& Multiple Frames  & 74.07 & 73.33 & 76.19 & 56.67 & 59.09 & 66.03 & 6.28 \\
\midrule
\multirow{2}{*}{\textbf{Claude-Sonnet-4.5}} 
& Single Frame         & 59.26 & 65.00 & 71.43 & 64.44 & 60.61 & 63.78 & 16.55 \\
& Multiple Frames  & 72.22 & 71.67 & 80.95 & 67.78 & 56.06 & 68.59 & 18.01 \\
\midrule
\multirow{2}{*}{\textbf{Claude-Haiku-4.5}} 
& Single Frame         & 50.00 & 55.00 & 64.29 & 48.89 & 56.06 & 53.85 & 7.14 \\
& Multiple Frames  & 53.70 & 58.33 & 64.29 & 55.56 & 60.61 & 58.01 & 8.46 \\
\midrule
\multirow{2}{*}{\textbf{Qwen-2.5-VL-72B}} 
& Single Frame         & 55.56 & 70.00 & 50.00 & 62.22 & 59.09 & 60.26 & 13.51 \\
& Multiple Frames  & 72.22 & 76.67 & 54.76 & 58.89 & 59.09 & 64.10 & 14.93 \\
\midrule
\multirow{2}{*}{\textbf{Qwen-2.5-VL-7B}} 
& Single Frame         & 53.70 & 61.67 & 61.90 & 57.78 & 56.06 & 58.01 & \textbf{4.14} \\
& Multiple Frames  & 72.22 & 66.67 & 59.52 & 54.44 & 46.97 & 58.97 & 4.29 \\
\midrule
\multirow{2}{*}{\textbf{OCR-Text Only GPT-4o}} 
& Single Frame         & 62.96 & 58.33 & 54.76 & 51.11 & 57.58 & 56.41 & 4.58 \\
& Multiple Frames  & 48.15 & 63.33 & 45.24 & 54.44 & 59.09 & 54.81 & 6.25 \\
% \midrule
% Feature Similarity       & N/A & 50.00 & 46.77 & 47.76 & 66.10 & 46.77 & 53.32 & \textbf{0.81} \\
\bottomrule
\end{tabular}
\vspace{-0.4cm}
\label{tab:main results}
\end{table*}

To examine both the efficiency and robustness of semantic reasoning in AR scenes, we designed two prompting strategies for inference:  
(1)~a single-frame setting, in which only one frame in the video is provided as input to simulate fast, real-time decision-making; and  
(2)~a multi-frame setting, in which multiple frames are provided to capture temporal context.  
In the single-frame setting, the frame selected corresponds to the exact midpoint of the video sequence, representing a typical view of the scene.  
In the multi-frame setting, three frames are sampled from each video: the first, middle, and last frames, providing the model with consecutive visual information across time. For Qwen models, due to their limited context size, the frames used in the multi-frame setting are downsampled by 50\%.
While this setting offers richer temporal cues, it also increases inference latency and, in some cases, may reduce accuracy due to the introduction of redundant or distracting information.  
Each model is prompted using a unified instruction template designed to elicit reasoning about semantic contradiction in AR content, as shown below. The words in brackets (e.g., image/images) indicate the differences between the two prompting strategies, where the singular form corresponds to the single-frame setting and the plural form corresponds to the multi-frame setting.

\begin{quote}
\footnotesize
\vspace{-0.2cm}
\leftskip=-1.5em
\rightskip=-1.5em
\textit{``You are an expert in evaluating augmented reality (AR) experiences.
        I will show you an image that represents the user's view of an AR application.
        Your task is to analyze whether the virtual content in the image is contradictory with each other. Please answer the following questions one by one:\\
        \textbf{1.} What is the real-world scenario shown in the image (\textit{images})? Provide a brief description of the environment and the objects present. \\
        \textbf{2.} Describe the virtual content in the image (\textit{images}). What are the virtual objects, where are they placed, and how do they visually interact with the real-world scene?\\
        \textbf{3.} Does the information of the virtual content in the image (\textit{images}) contradict with each other? Provide a detailed explanation of any inconsistencies or manipulations you identify. You should analyze whether the information provided by the virtual content is confusing or misleading to the user.\\
        \textbf{4.} Based on your analysis in Question 3, could the virtual content in this image (\textit{these images}) potentially lead to misunderstanding, distraction, or harm? If yes, answer only with **``Yes"**. Otherwise, answer with **``No"**."}
        
\vspace{-0.2cm}
\end{quote}

The commercial VLMs are accessed via their provider APIs, while open-source models are evaluated using the HuggingFace Inference API. After each model generates a response, we apply a post-processing step to extract the final prediction. Specifically, we examine the response and determine whether the answer to Question 4 in the prompt is ``Yes'' or ``No.'' In our experiments, all models produced explicit Yes/No answers for all cases.
These terms appear in the prompts solely to provide an intuitive interpretation for VLMs and encourage them to think like human users.
%, so the post-processing step simply reads this final decision and does not introduce any additional ambiguity.
%This direct pattern-matching approach avoids additional inference models and ensures that the decision is determined solely by the queried VLM’s own output. 
The overall experiment process is shown in Fig.~\ref{fig:diag}.

It is worth noting that although video-based multimodal models are emerging, we do not employ them here because (1) most VLM APIs do not yet support direct video input, and (2) continuous video streaming is impractical for real-time AR attack detection, where on-device compute and memory resources are severely constrained.

\subsection{Results and Analysis}

Tab~\ref{tab:main results} summarizes the attack detection accuracy and latency of 11 VLMs under both single-frame and multi-frame prompting conditions. All reported results are averaged over three independent runs. Overall, the best-performing model is GPT-5, achieving an average accuracy of 88.14\% in the single-frame setting, followed by Gemini-2.5-Pro (83.97\%, single frame) and GPT-4.1 (84.62\%, multiple frames). In contrast, smaller or open-source light-weighted models show limited reasoning capability, such as Qwen-2.5-VL-7B, which achieves an accuracy of 58.97\% (single frame), and Claude-Haiku-4.5 only reaches an accuracy of 53.85\% (single frame). The OCR-based text-only baseline also performed poorly, achieving an accuracy of 56.41\%. These results indicate that both open-source models and text-only approaches are substantially less effective for this task. The weaker performance of lightweight open-source VLMs is likely due to their limited semantic reasoning ability, while text-only baseline is unable to capture contradictions that require visual reasoning ability.

When comparing the two prompting strategies, the effect of multi-frame input varies across models. Among all the models, GPT-4.1, GPT-4o, Grok-2-Vision, two Claude models, and two Qwen models benefit from multi-frame prompting, gaining up to 5\% in overall accuracy, while GPT-5, two  Gemini models, and Grok-4 experience performance degradation (up to 8\%). This mixed trend suggests that while additional temporal context can provide complementary cues for understanding the AR scenes, it may also introduce redundant or conflicting information that overwhelms the model’s reasoning process. In particular, some VLMs are optimized for static image reasoning rather than multi-image temporal inference, which explains why some models fail to integrate the additional frames effectively. 

In terms of latency, all the experiments were conducted and measured under a network of 500 Mbps on average. We observe that multi-frame prompting consistently increases inference time across all models because of longer prompts and the need to process multiple visual embeddings. This trade-off is critical in real AR applications, where decisions must be made as quickly as possible, since long attack detection delays can reduce the user's experience and make the detection result less effective~\cite{arlatency01, arlatency02}. For example, GPT-5 achieves the highest accuracy but incurs a latency exceeding 19~s, which is almost impractical for interactive AR scenarios. In contrast, Qwen-2.5-VL-7B completes inference in only 4.14~s but yields a much lower accuracy of only 58.01\%, highlighting the trade-off between reasoning depth and real-time responsiveness. In contrast, GPT-4o maintains a strong balance between accuracy and responsiveness, achieving 84.62\% accuracy with an average latency of about 7~s. This balanced performance suggests that, under current commercial constraints, models with moderate complexity and efficient multimodal processing pipelines may be more suitable for real-time AR reasoning tasks such as contradictory virtual content detection.

\begin{figure}
    \centering
    \includegraphics[width=1\linewidth]{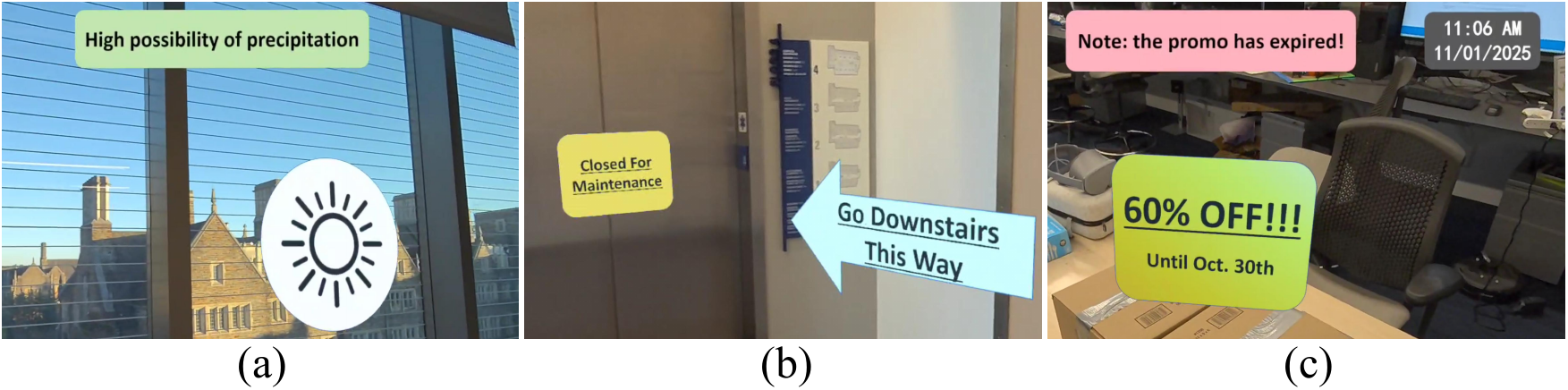}
    \vspace{-0.6cm}
    \caption{Three typical failure cases. Subfigure (a) and (b) are false negative cases and (c) is a false positive case.}
    \label{fig:failure}
    \vspace{-0.6cm}
\end{figure}

We also provide a qualitative failure analysis. We use GPT-5's results as examples and observe that there are mainly three reasons that could lead to VLM failures. 
First, VLM fails to recognize all the virtual content in the image.  In Fig.~\ref{fig:failure} (a), the sunny icon is not recognized as virtual content, and the recognized virtual content (text box) is not considered as contradiction. 
Second, VLM fails to understand virtual content in the context of real-world spatial information. In Fig.~\ref{fig:failure} (b), the arrow is leading the user to an elevator that is not working. However, while the VLM admits elevator is unavailable, it says ``directional arrow suggests a different alternate way," which is not true and leads to a prediction failure. 
Finally, VLM fails to reason the semantic meaning of virtual content. In Fig.~\ref{fig:failure} (c), the pop-up tells the user the promotion information in the virtual content has expired, according to the current time information. However, the VLM says ``the virtual elements conflict: one advertises an ongoing discount while the others state it is not valid." In fact, the discount is not necessarily an ongoing event.

These results collectively highlight that current VLMs possess a partial but limited capability to detect semantic contradictions in AR content, especially when low latency is required. While single-frame reasoning already yields competitive performance, achieving reliable multi-frame understanding without sacrificing efficiency remains an open challenge for next-generation multimodal models.

\vspace{-0.2cm}
\section{Limitations and Future Work}
\label{sec:limitation}

Although ContrAR represents the first systematic benchmark for evaluating VLMs under contradictory virtual content attacks, several limitations remain. First, the scale of the dataset can be further expanded. Our current dataset provides an initial benchmark foundation, but may not fully capture the diversity of AR environments or device rendering configurations. In future work, we plan to incorporate real-time attack generation using emerging 3D generative models~\cite{3dgeneration01, 3dgeneration02, 3dgeneration03} and automatic virtual content placement methods~\cite{xair, octo}, rather than relying solely on manually placing the pre-designed virtual content. Such dynamic and procedurally generated adversarial content may better approximate the nature of real-world AR attacks, enabling a more reliable and realistic benchmark.

Second, while the user-based validation confirms the reliability of the labels of the contradictory virtual content attack, their actual impact on user behavior has not yet been verified in real-world AR applications. The current validation captures offline judgments rather than the dynamic cognitive and behavioral responses that arise during immersive AR experience. Future work will therefore include conducting in-situ user studies on head-mounted AR devices. In addition, we will integrate VLM-based attack detection systems into the AR prototype during these studies, allowing us to assess not only how users respond to the attacks but also how they evaluate the detection system’s performance.

Finally, the current study focuses exclusively on semantic contradiction attacks, but the broader landscape of AR threats can be much wider. Future work will explore additional categories such as cognitive overload attacks~\cite{congitiveoverload01, congitiveoverload02}, which exploit users’ limited attention, and distraction attacks~\cite{distraction01}, where salient yet unimportant virtual content disrupts task performance without introducing explicit semantic conflict. Expanding the taxonomy and dataset to include these cases will enable a more comprehensive evaluation of AR security and user safety.

\vspace{-0.1cm}
\section{Conclusion}
\label{sec:conclusion}

In this paper, we introduce \textit{ContrAR}, the first benchmark for evaluating VLMs under contradictory virtual content attacks in AR environments. We formally define the attack through a comprehensive threat model, construct a real-world dataset with 312 user-validated AR videos, and conduct systematic evaluations across 11 state-of-the-art VLMs. Experiment results reveal that current models exhibit partial but limited capability in recognizing semantic contradictions among virtual elements, and that a clear trade-off exists between accuracy and inference latency. Our findings highlight both the promise and the current limitations of multimodal reasoning for AR safety. The benchmark provides a standardized, reproducible foundation for future research on secure and trustworthy AR perception. In future work, we aim to extend ContrAR with larger-scale datasets, additional attack types, and in-situ user studies to further explore how semantic and cognitive-level manipulations impact real-world AR experiences.

\section*{Acknowledgement}
We thank the participants of our user-based label validation for their invaluable effort and assistance in this research. This work was supported in part by NSF grants CSR-2312760, CNS-2112562, and IIS-2231975, NSF CAREER Award IIS-2046072, NSF NAIAD Award 2332744, a Cisco Research Award, a Meta Research Award, Defense Advanced Research Projects Agency Young Faculty Award HR0011-24-1-0001, and the Army Research Laboratory under Cooperative Agreement Number W911NF-23-2-0224. The views and conclusions contained in this document are those of the authors and should not be interpreted as representing the official policies, either expressed or implied, of the Defense Advanced Research Projects Agency, the Army Research Laboratory, or the U.S. Government. This paper has been approved for public release; distribution is unlimited. No official endorsement should be inferred. The U.S.~Government is authorized to reproduce and distribute reprints for Government purposes notwithstanding any copyright notation herein.

{
    \small
    \bibliographystyle{ieeenat_fullname}
    % \bibliography{main}

\begin{thebibliography}{45}
\providecommand{\natexlab}[1]{#1}
\providecommand{\url}[1]{\texttt{#1}}
\expandafter\ifx\csname urlstyle\endcsname\relax
  \providecommand{\doi}[1]{doi: #1}\else
  \providecommand{\doi}{doi: \begingroup \urlstyle{rm}\Url}\fi

\bibitem[Abid et~al.(2019)Abid, Abdalla, Abid, Khan, Alfozan, and Zou]{gradio}
Abubakar Abid, Ali Abdalla, Ali Abid, Dawood Khan, Abdulrahman Alfozan, and James Zou.
\newblock Gradio: Hassle-free sharing and testing of {ML} models in the wild.
\newblock \emph{arXiv preprint arXiv:1906.02569}, 2019.

\bibitem[Al-Kalbani et~al.(2019)Al-Kalbani, Frutos-Pascual, and Williams]{ARassessshadow01}
Maadh Al-Kalbani, Maite Frutos-Pascual, and Ian Williams.
\newblock Virtual object grasping in augmented reality: Drop shadows for improved interaction.
\newblock In \emph{Proceedings of International Conference on Virtual Worlds and Games for Serious Applications}, 2019.

\bibitem[{Anthropic}(2025{\natexlab{a}})]{anthropic2025claudehaiku45}
{Anthropic}.
\newblock Introducing the {C}laude {H}aiku 4.5.
\newblock \url{https://www.anthropic.com/news/claude-haiku-4-5}, 2025{\natexlab{a}}.

\bibitem[{Anthropic}(2025{\natexlab{b}})]{anthropic2025claudesonnet45}
{Anthropic}.
\newblock Introducing the {C}laude {S}onnet 4.5.
\newblock \url{https://www.anthropic.com/news/claude-sonnet-4-5}, 2025{\natexlab{b}}.

\bibitem[Buchner et~al.(2022)Buchner, Buntins, and Kerres]{congitiveoverload02}
Josef Buchner, Katja Buntins, and Michael Kerres.
\newblock The impact of augmented reality on cognitive load and performance: A systematic review.
\newblock \emph{Journal of Computer Assisted Learning}, 38\penalty0 (1), 2022.

\bibitem[Chen et~al.(2025{\natexlab{a}})Chen, Andreyev, Xiu, Chilukuri, Sen, Imani, Li, Gorlatova, Tan, and Lan]{chen2025neurosymbolic}
Rongqian Chen, Allison Andreyev, Yanming Xiu, Joshua Chilukuri, Shunav Sen, Mahdi Imani, Bin Li, Maria Gorlatova, Gang Tan, and Tian Lan.
\newblock A neurosymbolic framework for interpretable cognitive attack detection in augmented reality.
\newblock \emph{arXiv preprint arXiv:2508.09185}, 2025{\natexlab{a}}.

\bibitem[Chen et~al.(2025{\natexlab{b}})Chen, Hong, Islam, Imani, Tan, and Lan]{ARcognitiveattack}
Rongqian Chen, Shu Hong, Rifatul Islam, Mahdi Imani, Gang Tan, and Tian Lan.
\newblock Demo: Perception graph for cognitive attack reasoning in augmented reality.
\newblock In \emph{Proceedings of International Symposium on Theory, Algorithmic Foundations, and Protocol Design for Mobile Networks and Mobile Computing}, 2025{\natexlab{b}}.

\bibitem[Chen and Wu(2024)]{chen2024securing}
Si Chen and Jie Wu.
\newblock Securing augmented reality applications.
\newblock In \emph{Network Security Empowered by Artificial Intelligence}, pages 331--354. Springer, 2024.

\bibitem[Cheng et~al.(2023)Cheng, Tian, Kohno, and Roesner]{attack01}
Kaiming Cheng, Jeffery~F. Tian, Tadayoshi Kohno, and Franziska Roesner.
\newblock Exploring user reactions and mental models towards perceptual manipulation attacks in mixed reality.
\newblock In \emph{Proceedings of USENIX Security Symposium}, 2023.

\bibitem[De~Marneffe et~al.(2008)De~Marneffe, Rafferty, and Manning]{contra01}
Marie-Catherine De~Marneffe, Anna~N Rafferty, and Christopher~D Manning.
\newblock Finding contradictions in text.
\newblock In \emph{Proceedings of ACL-08: HLT}, 2008.

\bibitem[Doswell and Skinner(2014)]{congitiveoverload01}
Jayfus~T Doswell and Anna Skinner.
\newblock Augmenting human cognition with adaptive augmented reality.
\newblock In \emph{International Conference on Augmented Cognition}, 2014.

\bibitem[Duan et~al.(2022)Duan, Guo, Sun, Min, Chen, and Zhai]{ARIQA01}
Huiyu Duan, Lantu Guo, Wei Sun, Xiongkuo Min, Li Chen, and Guangtao Zhai.
\newblock Augmented reality image quality assessment based on visual confusion theory.
\newblock In \emph{Proceedings of IEEE International Symposium on Broadband Multimedia Systems and Broadcasting (BMSB)}, 2022.

\bibitem[Duan et~al.(2025{\natexlab{a}})Duan, Rotondo, Xiu, Eom, Chen, Li, Hu, and Gorlatova]{DiverseARplus}
Lin Duan, Elias Rotondo, Yanming Xiu, Sangjun Eom, Ryan Chen, Conrad Li, Yuhe Hu, and Maria Gorlatova.
\newblock Probing the augmented reality scene analysis capabilities of large multimodal models: Toward reliable real-time assessment solutions.
\newblock \emph{IEEE Internet Computing}, 2025{\natexlab{a}}.

\bibitem[Duan et~al.(2025{\natexlab{b}})Duan, Xiu, and Gorlatova]{Genaixrws}
Lin Duan, Yanming Xiu, and Maria Gorlatova.
\newblock Advancing the understanding and evaluation of {AR}-generated scenes: When vision-language models shine and stumble.
\newblock In \emph{Proceedings of IEEE Conference on Virtual Reality and 3D User Interfaces Abstracts and Workshops (VRW)}, 2025{\natexlab{b}}.

\bibitem[et~al.(2025)]{gemini2.5}
Gheorghe~Comanici et al.
\newblock Gemini 2.5: Pushing the frontier with advanced reasoning, multimodality, long context, and next generation agentic capabilities, 2025.

\bibitem[Fu et~al.(2022)Fu, Fang, Gao, Hong, and Chen]{sceneunderstanding03}
Mingyu Fu, Wei Fang, Shan Gao, Jianhao Hong, and Yizhou Chen.
\newblock Edge computing-driven scene-aware intelligent augmented reality assembly.
\newblock \emph{The International Journal of Advanced Manufacturing Technology}, 119\penalty0 (11), 2022.

\bibitem[{Gemini Team}(2024)]{geminiteam2024geminifamilyhighlycapable}
{Gemini Team}.
\newblock Gemini: A family of highly capable multimodal models.
\newblock \emph{arXiv:2312.11805}, 2024.

\bibitem[Guti{\'e}rrez et~al.(2020)Guti{\'e}rrez, Vigier, and Le~Callet]{ARassessmentlighting01}
Jes{\'u}s Guti{\'e}rrez, Toinon Vigier, and Patrick Le~Callet.
\newblock Quality evaluation of {3D} objects in mixed reality for different lighting conditions.
\newblock \emph{Electronic Imaging}, 32, 2020.

\bibitem[Kazemi et~al.(2023)Kazemi, Yuan, Bhatia, Kim, Xu, Imbrasaite, and Ramachandran]{contradiction01}
Mehran Kazemi, Quan Yuan, Deepti Bhatia, Najoung Kim, Xin Xu, Vaiva Imbrasaite, and Deepak Ramachandran.
\newblock {BoardgameQA}: A dataset for natural language reasoning with contradictory information.
\newblock In \emph{Advances in Neural Information Processing Systems}, 2023.

\bibitem[Kim and Gabbard(2022)]{distraction01}
Hyungil Kim and Joseph~L Gabbard.
\newblock Assessing distraction potential of augmented reality head-up displays for vehicle drivers.
\newblock \emph{Human Factors}, 64\penalty0 (5), 2022.

\bibitem[Kim et~al.(2011)Kim, Kim, and Lee]{ARassessphysics01}
Sinyoung Kim, Yeonjoon Kim, and Sunghee Lee.
\newblock On visual artifacts of physics simulation in augmented reality environment.
\newblock In \emph{Proceedings of International Symposium on Ubiquitous Virtual Reality}, 2011.

\bibitem[Lebeck et~al.(2018)Lebeck, Ruth, Kohno, and Roesner]{attack02}
K. Lebeck, K. Ruth, T. Kohno, and F. Roesner.
\newblock Arya: Operating system support for securely augmenting reality.
\newblock \emph{IEEE Security \& Privacy}, 16\penalty0 (01), 2018.

\bibitem[Lee et~al.(2010)Lee, Bonebrake, Bowman, and H{\"o}llerer]{arlatency01}
Cha Lee, Scott Bonebrake, Doug~A Bowman, and Tobias H{\"o}llerer.
\newblock The role of latency in the validity of {AR} simulation.
\newblock In \emph{Proceedings of IEEE Virtual Reality Conference (VR)}, 2010.

\bibitem[Li et~al.(2024{\natexlab{a}})Li, Suzuki, Ohtake, Yatagawa, and Matsuda]{ARassessmisalign01}
Tinghao Li, Hiromasa Suzuki, Yutaka Ohtake, Tatsuya Yatagawa, and Shinji Matsuda.
\newblock Efficient evaluation of misalignment between real and virtual objects for {HMD}-based {AR} assembly assistance system.
\newblock \emph{Advanced Engineering Informatics}, 59, 2024{\natexlab{a}}.

\bibitem[Li et~al.(2024{\natexlab{b}})Li, Zhang, Kang, Cheng, Gao, Zhang, Liang, Liao, Cao, and Shan]{3dgeneration01}
Xiaoyu Li, Qi Zhang, Di Kang, Weihao Cheng, Yiming Gao, Jingbo Zhang, Zhihao Liang, Jing Liao, Yanpei Cao, and Ying Shan.
\newblock Advances in 3{D} generation: A survey.
\newblock \emph{arXiv preprint arXiv:2401.17807}, 2024{\natexlab{b}}.

\bibitem[Likert(1932)]{likert}
Rensis Likert.
\newblock A technique for the measurement of attitudes.
\newblock \emph{Archives of {P}sychology}, 1932.

\bibitem[OpenAI(2023)]{GPT4V}
OpenAI.
\newblock {GPT-4} technical report.
\newblock \emph{arXiv:2303.08774}, 2023.

\bibitem[OpenAI(2025)]{GPT5}
OpenAI.
\newblock {GPT-5} system card.
\newblock \emph{https://cdn.openai.com/gpt-5-system-card.pdf}, 2025.

\bibitem[Pan et~al.(2023)Pan, Chen, Kan, and Wang]{contradiction02}
Liangming Pan, Wenhu Chen, Min-Yen Kan, and William~Yang Wang.
\newblock Attacking open-domain question answering by injecting misinformation.
\newblock In \emph{Proceedings of the 13th International Joint Conference on Natural Language Processing and the 3rd Conference of the Asia-Pacific Chapter of the Association for Computational Linguistics (IJCNLP-AACL)}, 2023.

\bibitem[Pauly et~al.(2015)Pauly, Diotte, Fallavollita, Weidert, Euler, and Navab]{sceneunderstanding02}
Olivier Pauly, Benoit Diotte, Pascal Fallavollita, Simon Weidert, Ekkehard Euler, and Nassir Navab.
\newblock Machine learning-based augmented reality for improved surgical scene understanding.
\newblock \emph{Computerized Medical Imaging and Graphics}, 41, 2015.

\bibitem[Pedziwiatr et~al.(2022)Pedziwiatr, K{\"u}mmerer, Wallis, Bethge, and Teufel]{contra02}
Marek~A Pedziwiatr, Matthias K{\"u}mmerer, Thomas~SA Wallis, Matthias Bethge, and Christoph Teufel.
\newblock Semantic object-scene inconsistencies affect eye movements, but not in the way predicted by contextualized meaning maps.
\newblock \emph{Journal of Vision}, 22\penalty0 (2), 2022.

\bibitem[Sharma et~al.(2024)Sharma, Yoffe, and H\"ollerer]{octo}
Aditya Sharma, Luke Yoffe, and Tobias H\"ollerer.
\newblock {{OCTO+}: A Suite for Automatic Open-Vocabulary Object Placement in Mixed Reality}.
\newblock In \emph{Proceedings of IEEE International Conference on Artificial Intelligence and eXtended and Virtual Reality (AIxVR)}, 2024.

\bibitem[Slocum et~al.(2024)Slocum, Zhang, Shayegani, Zaree, Abu-Ghazaleh, and Chen]{VIM01}
Carter Slocum, Yicheng Zhang, Erfan Shayegani, Pedram Zaree, Nael Abu-Ghazaleh, and Jiasi Chen.
\newblock That doesn{\textquoteright}t go there: Attacks on shared state in multi-user augmented reality applications.
\newblock In \emph{Proceedings of USENIX Security Symposium}, 2024.

\bibitem[Sprute et~al.(2019)Sprute, Viertel, Tönnies, and König]{sceneunderstanding01}
Dennis Sprute, Philipp Viertel, Klaus Tönnies, and Matthias König.
\newblock Learning virtual borders through semantic scene understanding and augmented reality.
\newblock In \emph{Proceedings of IEEE/RSJ International Conference on Intelligent Robots and Systems (IROS)}, 2019.

\bibitem[Srinidhi et~al.(2024)Srinidhi, Lu, and Rowe]{xair}
Sruti Srinidhi, Edward Lu, and Anthony Rowe.
\newblock {XaiR}: An {XR} platform that integrates large language models with the physical world.
\newblock In \emph{Proceedings of IEEE International Symposium on Mixed and Augmented Reality (ISMAR)}, 2024.

\bibitem[Tochilkin et~al.(2024)Tochilkin, Pankratz, Liu, Huang, Letts, Li, Liang, Laforte, Jampani, and Cao]{3dgeneration02}
Dmitry Tochilkin, David Pankratz, Zexiang Liu, Zixuan Huang, Adam Letts, Yangguang Li, Ding Liang, Christian Laforte, Varun Jampani, and Yanpei Cao.
\newblock Tripo{SR}: Fast 3{D} object reconstruction from a single image.
\newblock \emph{arXiv preprint arXiv:2403.02151}, 2024.

\bibitem[Tseng et~al.(2022)Tseng, Bonnail, McGill, Khamis, Lecolinet, Huron, and Gugenheimer]{VIM03}
Wenjie Tseng, Elise Bonnail, Mark McGill, Mohamed Khamis, Eric Lecolinet, Samuel Huron, and Jan Gugenheimer.
\newblock The dark side of perceptual manipulations in virtual reality.
\newblock In \emph{Proceedings of the ACM CHI Conference on Human Factors in Computing Systems}, 2022.

\bibitem[Wasenm{\"u}ller et~al.(2016)Wasenm{\"u}ller, Meyer, and Stricker]{ARassessmisalign02}
Oliver Wasenm{\"u}ller, Marcel Meyer, and Didier Stricker.
\newblock Augmented reality 3{D} discrepancy check in industrial applications.
\newblock In \emph{Proceedings of IEEE International Symposium on Mixed and Augmented Reality (ISMAR)}, 2016.

\bibitem[XAI(2025)]{Grok4}
XAI.
\newblock Grok 4 fast model card.
\newblock \emph{https://data.x.ai/2025-09-19-grok-4-fast-model-card.pdf}, 2025.

\bibitem[Xiang et~al.(2025)Xiang, Lv, Xu, Deng, Wang, Zhang, Chen, Tong, and Yang]{3dgeneration03}
Jianfeng Xiang, Zelong Lv, Sicheng Xu, Yu Deng, Ruicheng Wang, Bowen Zhang, Dong Chen, Xin Tong, and Jiaolong Yang.
\newblock Structured 3{D} latents for scalable and versatile 3d generation.
\newblock In \emph{Proceedings of the IEEE/CVF Computer Vision and Pattern Recognition Conference (CVPR)}, 2025.

\bibitem[Xiu and Gorlatova(2025{\natexlab{a}})]{vimsense}
Yanming Xiu and Maria Gorlatova.
\newblock Detecting visual information manipulation attacks in augmented reality: A multimodal semantic reasoning approach.
\newblock \emph{IEEE Transactions on Visualization and Computer Graphics}, 31\penalty0 (11), 2025{\natexlab{a}}.

\bibitem[Xiu and Gorlatova(2025{\natexlab{b}})]{xiu2025demonstrating}
Yanming Xiu and Maria Gorlatova.
\newblock Demonstrating visual information manipulation attacks in augmented reality: A hands-on miniature city-based setup.
\newblock In \emph{Proceedings of the International Symposium on Theory, Algorithmic Foundations, and Protocol Design for Mobile Networks and Mobile Computing (MobiHoc)}, 2025{\natexlab{b}}.

\bibitem[Xiu et~al.(2025)Xiu, Scargill, and Gorlatova]{viddar}
Yanming Xiu, Tim Scargill, and Maria Gorlatova.
\newblock { {ViDDAR}: Vision Language Model-Based Task-Detrimental Content Detection for Augmented Reality }.
\newblock \emph{IEEE Transactions on Visualization and Computer Graphics}, 31\penalty0 (05), 2025.

\bibitem[Yan et~al.(2025)Yan, Fan, Li, Jiang, Zhao, Guan, Kuo, and Wang]{multimodalinconsistency}
Qianqi Yan, Yue Fan, Hongquan Li, Shan Jiang, Yang Zhao, Xinze Guan, Ching-Chen Kuo, and Xin~Eric Wang.
\newblock Multimodal inconsistency reasoning ({MMIR}): A new benchmark for multimodal reasoning models.
\newblock In \emph{Findings of the Association for Computational Linguistics (ACL)}, 2025.

\bibitem[Zhang et~al.(2022)Zhang, Han, and Hui]{arlatency02}
Wenxiao Zhang, Bo Han, and Pan Hui.
\newblock Sear: Scaling experiences in multi-user augmented reality.
\newblock \emph{IEEE Transactions on Visualization and Computer Graphics}, 28\penalty0 (5), 2022.

\end{thebibliography}

}


\end{document}